\documentclass{article}
\usepackage{spconf,amsmath,graphicx}
\usepackage{xcolor}
\usepackage{booktabs}
\usepackage[normalem]{ulem}
\useunder{\uline}{\ul}{}
\usepackage{subcaption}
\usepackage{amssymb}

\title{3D Human Pose Regression using Graph Convolutional Network}

\name{Soubarna Banik$^{\star}$ \qquad Alejandro Mendoza Garc\'{i}a$^{\dagger}$ \qquad Alois Knoll$^{\star}$}

\address{$^{\star}$ Technical University of Munich, Department of Informatics, Munich, Germany  \\
    $^{\dagger}$ reFit Systems, Munich, Germany
    }

\usepackage{tikz}

\begin{document}
%\ninept
%
\maketitle
%\blfootnote{\copyrightnotice}

%
\begin{abstract}
3D human pose estimation is a difficult task, due to challenges such as occluded body parts and ambiguous poses. Graph convolutional networks encode the structural information of the human skeleton in the form of an adjacency matrix, which is beneficial for better pose prediction. We propose one such graph convolutional network named PoseGraphNet for 3D human pose regression from 2D poses. Our network uses an adaptive adjacency matrix and kernels specific to neighbor groups. We evaluate our model on the Human3.6M dataset which is a standard dataset for 3D pose estimation.
Our model's performance is close to the state-of-the-art, but with much fewer parameters.
The model learns interesting adjacency relations between joints that have no physical connections, but are behaviorally similar.
\end{abstract}
\begin{keywords}
3D Human pose regression, Graph Convolutional Networks, CNN
\end{keywords}
\section{Introduction}
\label{sec:intro}

3D human pose estimation (HPE) directly from RGB images is a challenging task.
Compared to RGB images 3D point cloud data has inherent depth information, which is useful for 3D pose prediction but is computationally expensive to work with \cite{Moon2018,Zimmermann2018}.
Recent approaches \cite{Tome2017,Rayat2018,Cai2019} have split the task into two parts: predicting 2d pose from an image, and then lifting them to 3D poses.
Restricting the 3D pose to the camera space allowed the 2D to 3D pose task to be learnable using only supervised learning techniques~\cite{Martinez2017}.

It has been shown that additional information such as structural constraints of the skeleton improves the performance further~\cite{marin20183d,Cai2019}.
In some earlier work, the structural information is provided to the system by a separate statistical model \cite{Tome2017} or by a dictionary of basic poses \cite{marin20183d}.
These types of structural models need to be learned separately. 
As the learned structural model is dependent on the training dataset, it also restricts the number of possible 3D poses.

Recent advances in graph convolutional networks (GCN) and their use in pose estimation look promising \cite{Cai2019,Zhao2019,Doosti2020}.
GCNs provide a lightweight solution for representing the structure of a human body without any additional overhead.
Zhao et al.~\cite{Zhao2019} propose a semantic GCN for encoding both local and global relationships among different body joints for 2D to 3D pose regression.
Cai et al.~\cite{Cai2019} use a spatio-temporal GCN to capture the temporal constraints in addition to the structural constraints. They also capture multi-scale features by using a U-Net architecture with residual connections.
Hope-Net~\cite{Doosti2020} uses an adaptive adjacency matrix and a graph U-Net architecture for hand-object pose estimation.

In this paper, we propose a graph convolutional network named PoseGraphNet for 2D to 3D pose regression.
Our model is similar to \cite{Martinez2017}, but uses GCN instead of a fully connected network.
It uses adaptive adjacency matrices and kernels specific to neighbor groups.
Our method outperforms the base model \cite{Martinez2017} on  Human3.6m dataset \cite{ionescu2013human3} and is close to the performance of the state-of-the-art graph-based approaches \cite{Cai2019} despite having much fewer parameters.

\begin{figure*}
\begin{center}
%\fbox{\rule{0pt}{2in} \rule{.9\linewidth}{0pt}}
\includegraphics[width=0.9\textwidth]{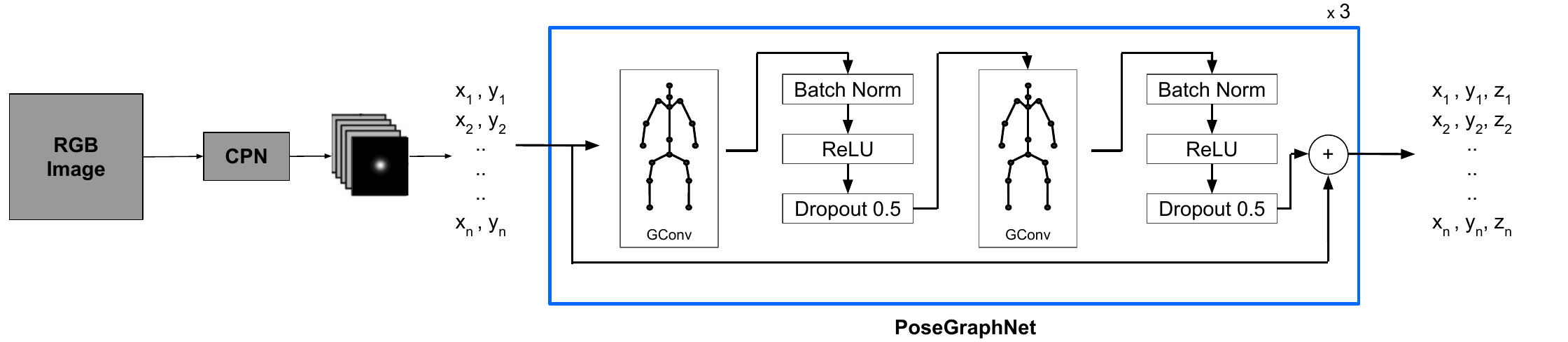}
\end{center}
   \caption{Overview of the 3D pose regression algorithm. The proposed PoseGraphNet is highlighted by the blue rectangle.}
\label{fig:pgn}
\end{figure*}

\section{PoseGraphNet}

We propose a graph convolution-based network, named PoseGraphNet for 3D pose regression. 
Figure \ref{fig:pgn} describes the overall pipeline and the architecture of our proposed PoseGraphNet model.
We use a baseline 2D pose detection model named Cascaded Pyramid Network (CPN) \cite{Chen2018} which takes an RGB image as input and generates a heatmap for every joint, indicating the location of the joint in the image.
We extract the 2D joint coordinates from the heatmaps and pass them to our PoseGraphNet model.
The output of our model is the 3D position of the joints relative to the root joint.
The input joint coordinates are in image space and the outputs are in camera coordinate space. 

\textbf{Graph Convolutional Networks:} GCN is a neural network that works directly on graphs and can learn features of graph-structured data.
In graph-based pose estimation, the human skeleton is represented as a graph, $G=(V,E)$, where $V$ denotes the set of $N$ body joints %, $v_i \in V$ 
and $E$ denotes the set of edges signifying the connections between the body joints. %, $(v_i,v_j) \in E$.
The edges are represented by means of an adjacency matrix, $A \in \mathbb{R}^{N \times N}$.
The graph consists of node-level features and $A$ indicates, how the features of a node's neighbors and itself will be aggregated to form the output of the layer.

\subsection{Architecture}
\label{sec:archi}
Our proposed PoseGraphNet is based on \cite{Martinez2017} by Martinez et al. 
We choose this model in comparison to the more recent, U-shaped graph-based models~\cite{Cai2019,Doosti2020}, as the feed-forward network of \cite{Martinez2017} has less computational complexity and performs well without utilizing any structural information of the skeleton. 
We also want to find out how much the structural information and graph convolution alone help in improving the performance compared to that of \cite{Martinez2017}.
PoseGraphNet consists of 3 residual blocks, where each block is made of two units of a graph convolution layer with $768$ feature channels, batch normalization, a ReLU activation, and dropout. %, as depicted in Figure \ref{fig:pgn}. 
The block input is added to the block output with a residual connection.
An initial unit of graph convolution of $768$ channels and a final graph convolution layer of 3 output channels are not depicted in Figure \ref{fig:pgn} due to space constraints.

\textbf{Adaptive adjacency matrix}: In addition to learning the trainable weights of the graph convolution layers, we also learn the adjacency matrix as in \cite{Doosti2020}.
A pre-defined adjacency matrix $A$ representing the default physical structure of the human body
allows only equal contribution by every neighbor.
We observe, for most actions some body joints such as the trunk are relatively stable and do not move rapidly compared to the joints at the extremities.
Therefore, a neighbor joint situated at a body extremity should have a different contribution compared to a neighbor joint that is closer to the stable body trunk.
An adaptive adjacency matrix allows learning different contribution rates for different types of neighbors in addition to learning physically missing but behaviorally similar connections.

\textbf{Kernels for different neighbor groups}: In addition to the adaptive adjacency matrix, we learn different weights for different types of neighbors following~\cite{Cai2019}, resulting in the extraction of distinct features for different neighbor groups.
We categorize the neighboring nodes into three groups - a) self, b) parent (closer to the pelvis joint), and c) child (away from the pelvis joint). 
Output features of hidden layer $i$ is computed as,
\begin{align}
H_i & = \sigma(\tilde{A}^{self}H_{i-1}W^{self}_{i} + \tilde{A}^{parent}H_{i-1}W^{parent}_{i} + \nonumber \\
&\qquad\tilde{A}^{child}H_{i-1}W^{child}_{i}),
\label{eq:split_kernels}
\end{align}
where $\sigma$ is the activation function, $H_{i-1} \in \mathbb{R}^{N \times F_{i-1}}$ is the input feature matrix, $F_i$ is the number of features in layer $i$ and $W_i^{self},W_i^{parent},W_i^{child} \in \mathbb{R}^{F_{i-1} \times F_{i}}$ are the trainable weight matrices of layer $i$, for respective neighbor groups.
$\tilde{A}^{self}$, $\tilde{A}^{parent}$ and $\tilde{A}^{child}$ are the adjaceny matrices of the respective neighbor groups, normalized symmetrically with their corresponding degree matrices.
Note, all three adjacency matrices are adaptive and learned together with the weight matrices. 

\begin{table*}[t]
\centering
\resizebox{\textwidth}{!}{\begin{tabular}{lllllllllllllllll}
\toprule
%RHip & LHip & Spine & Thorax & Neck & LSho. & RSho. & LKnee & RKnee & Head & LElb. & RFoot & RElb. & LFoot & RWrist & LWrist & Avg. \\
RHip & LHip & Spine & Thorax & Neck & LSho. & RKnee & LKnee &  RSho.& Head & LElb. & RFoot & RElb. & LFoot & RWrist & LWrist & Avg. \\
\hline
%22.3 & 23.5 & 35.6  & 43.9   & 53.6 & 55.0  & 55.2  & 56.7  & 57.3  & 60.3 & 79.5  & 79.6  & 83.0  & 91.6  & 106.9  & 108.1  & 59.5 \\
22.0 & 22.5 & 34.6  & 38.6   & 48.1 & 48.3  &  48.8 & 49.0 & 49.2 & 52.7 & 69.7 & 71.9 & 72.2 & 81.4 & 93.6 & 94.3 & 52.8 \\
\bottomrule
\end{tabular}
}
\caption{\label{tab:mpjpe-joint}Mean Per Joint Position Error (MPJPE) (Protocol \#1) in millimeter between predicted and ground-truth poses on Human 3.~6m Dataset at joint level, in increasing order. The last column reports the average error. }
\end{table*}

\begin{table*}[t]
  \centering
\resizebox{\textwidth}{!}{\begin{tabular}{lllllllllllllllll}
\toprule
\textbf{Protocol \#1}                                        & Dir.  & Disc. & Eat   & Greet & Phone & Photo & Pose  & Purch. & Sit   & SitD. & Smoke & Wait  & WalkD. & Walk  & WalkT. & Avg.  \\
\hline
Martinez et al. \cite{Martinez2017} ICCV‘17 & 51.8  & 56.2  & 58.1  & 59.0  & 69.5  & 78.4  & 55.2  & 58.1   & 74.0  & 94.6  & 62.3  & 59.1  & 65.1   & 49.5  & 52.4   & 62.9  \\
Cai et al. \cite{Cai2019} ICCV‘19           & {46.5}  & \textbf{48.8}  & {47.6}  & {50.9}  & \textbf{52.9}  & 61.3  & 48.3  & \textbf{45.8 }  & \textbf{59.2}  & 64.4  & \textbf{51.2}  & \textbf{48.4}  & 53.5   & \textbf{39.2}  & {41.2}   & \textbf{50.6}  \\
Zhao et al. \cite{Zhao2019} CVPR‘19         & 47.3  & 60.7  & 51.4  & 60.5  & 61.1  & \textbf{49.9}  & \textbf{47.3}  & 68.1   & 86.2  & \textbf{55.0}  & 67.8  & 61.0  & \textbf{42.1}   & 60.6  & 45.3   & 57.6  \\
Pavllo et al \cite{Pavllo2019} CVPR‘19      & 47.1  & 50.6  & 49.0  & 51.8  & 53.6  & 61.4  & 49.4  & 47.4   & 59.3  & 67.4  & 52.4  & 49.5  & 55.3   & 39.5  & 42.7   & 51.8  \\
Pavlakos et al. \cite{Pavlakos2018} CVPR‘18 & 48.5  & 54.4  & 54.4  & 52.0  & 59.4  & 65.3  & 49.9  & 52.9   & 65.8  & 71.1  & 56.6  & 52.9  & 60.9   & 44.7  & 47.8   & 56.2  \\
\hline
Ours  (Mask-RCNN)                                                       & 51.0         & 55.3         & 54.0         & 54.6         & 62.4         & 76.0         & 51.6         & 52.7         & 79.3         & 87.1         & 58.4         & 56.0         & 61.8         & 48.1         & 44.1         & 59.5         \\
Ours (CPN) & \textbf{44.8} & 51.0 & \textbf{47.2} & \textbf{50.4} & 54.5 & 64.6 & 50.1 & 48.4 & 67.2 & 72.2 & 51.6 & 50.1 & 54.8 & 43.6 & \textbf{39.5} & 52.8 \\
\toprule
\textbf{Protocol \#2}                                        & Dir.  & Disc. & Eat   & Greet & Phone & Photo & Pose  & Purch. & Sit   & SitD. & Smoke & Wait  & WalkD. & Walk  & WalkT. & Avg.  \\
\hline
Martinez et al. \cite{Martinez2017} ICCV‘17 & 39.5  & 43.2  & 46.4  & 47.0  & 51.0  & 56.0  & 41.4  & 40.6   & 56.5  & 69.4  & 49.2  & 45.0  & 49.5   & 38.0  & 43.1   & 47.7  \\
Cai et al. \cite{Cai2019} ICCV‘19           & 36.8  & \textbf{38.7}  & 38.2  & 41.7  & 40.7  & 46.8  & 37.9  & 35.6   & 47.6  & \textbf{51.7}  & \textbf{41.3}  & \textbf{36.8}  & \textbf{42.7}   & 31.0  & {34.7}   & 40.2  \\
Pavllo et al \cite{Pavllo2019} CVPR‘19      & 36.0  & \textbf{38.7}  & \textbf{38.0}  & 41.7  & \textbf{40.1}  & \textbf{45.9}  & \textbf{37.1}  & \textbf{35.4}   & \textbf{46.8}  & 53.4  & 41.4  & 36.9  & 43.1   & \textbf{30.3}  & 34.8   & \textbf{40.0}  \\
Pavlakos et al. \cite{Pavlakos2018} CVPR‘18 & \textbf{34.7}  & 39.8  & 41.8  & \textbf{38.6}  & 42.5  & 47.5  & 38.0  & 36.6   & 50.7  & 56.8  & 42.6  & 39.6  & 43.9   & 32.1  & 36.5   & 41.8  \\
\hline
Ours  (Mask-RCNN)                                                      & 38.4         & 43.1         & 42.9         & 44.0          & 47.8         & 56.0         & 39.3         & 39.8         & 61.8         & 67.1         & 46.1        & 43.4         & 48.4         & 40.7         & 35.1         & 46.4        \\
Ours (CPN) & 35.0 & 40.0 & 38.1 & 41.4 & 42.1 & 48.6 & 38.4 & 37.0 & 53.1 & 57.5 & 41.5 & 38.2 & 43.6 & 37.6 & \textbf{31.4} & 41.6\\
\bottomrule
\end{tabular}
}
\caption{\label{tab:mpjpe}Mean Per Joint Position Error (MPJPE) in millimeter between predicted and ground-truth 3D poses on Human 3.6m Dataset under Protocol \#1 and Protocol \#2. The lower the error, the better. The best scores are highlighted in bold.}
\end{table*}

\begin{table}[]
\centering
\resizebox{\linewidth}{!}{
\begin{tabular}{ccrc}
\toprule
Model & Parameters & Throughput & MPJPE (mm) \\
\hline
Cai et al.~\cite{Cai2019} & 3.77M & 95.53 & 50.6 \\
Martinez et al.~\cite{Martinez2017} & 4.29M & 911.92 & 62.9 \\
Ours & 1.25M & 207.29 & 52.8 \\
\bottomrule
\end{tabular}
}
\caption{\label{tab:num_param}Number of parameters, Throughput (forward passes/second with batch size 1), and MPJPE (Protocol-1) of models. Higher throughput is better.}
\end{table}

%-------------------------------------------------------------------------
\section{Results}

\subsection{Implementation Details}
In our experiment, the 2D inputs are in the image coordinate system and are normalized so that image width is in the range $[-1,+1]$, maintaining the image aspect ratio.
For the 3D outputs, we use the camera coordinate system, instead of an arbitrary world coordinate system.
As pointed out in \cite{Martinez2017,Cai2019,Zhao2019}, it is unrealistic to train a model to infer 3D joint locations in an arbitrary coordinate space, as there can be multiple possible solutions for the same 2D input.
The camera frame makes the 2D to 3D joint prediction task coherent across different camera views.
Also, we are interested in the relative positions of the joints with respect to the root joint, which is the pelvis joint.
Hence, we zero-center the output joints by subtracting the root joint.

\subsection{Training Details}
We train our model using the Adam optimizer \cite{KingmaB14} for 50 epochs, using a batch size of $256$, a dropout rate of  $0.20$.
We start with a learning rate of $0.0001$, and decay it by 0.92 after every 10 epochs.
The network is initialized using Xavier initialization \cite{glorot2010understanding}. 
We do not use any data augmentation.
We minimize the mean squared error (MSE) loss between the predicted joints and the ground truth joint, $L(J^{pred}, J^{gt}) = \sum_{i=1}^{N} ||J^{pred}_i - J^{gt}_i||^2,$ where $J^{pred}$ and $J^{gt}$ are the predicted and ground truth 3D joints respectively.
The hip joint is excluded from evaluation as it remains zero throughout.

\subsection{Dataset}
We evaluate our proposed approach on the Human3.6M dataset \cite{ionescu2013human3}, a standard dataset for 3D human pose estimation.
The dataset contains 3.6 million RGB images of 7 subjects, from 4 different viewpoints performing 15 actions such as walking, eating, sitting, discussing, etc.
2D locations in image space and 3D ground truth positions in camera space of 17 body joints are available.
As per the standard protocol, we use images of all 4 viewpoints of subjects 1, 5, 6, 7, and 8 for training, and subjects 9 and 11 for evaluation.

\subsection{Evaluation Metrics}
We evaluate the PoseGraphNet using mean per joint position error (MPJPE), which computes the mean Euclidean distance of the predicted 3D joints to the ground truth joints in millimeters.
In previous work MPJPE is computed under two protocols~\cite{Martinez2017,Cai2019,Zhao2019}.
As per \textit{Protocol \#1}, the error is computed after aligning the root joints. 
In \textit{Protocol \#2} the 3D pose is further aligned with the ground truth using a rigid transformation (Procrustes alignment), before computing the error.

\begin{figure*}[t]
\begin{center}
%\fbox{\rule{0pt}{2in} \rule{.9\linewidth}{0pt}}
\includegraphics[width=\textwidth]{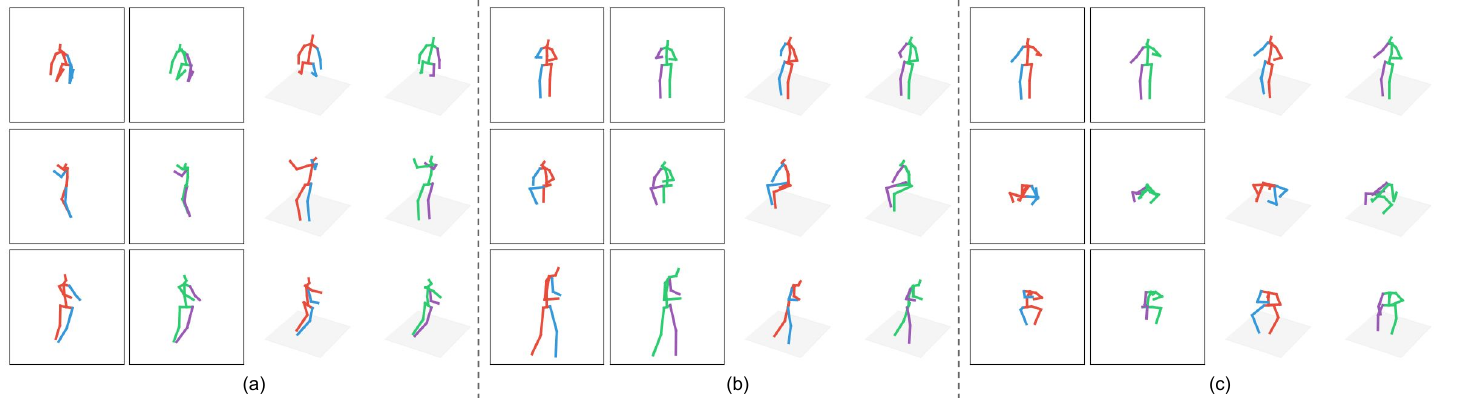}
\end{center}
   \caption{Qualitative results on the test set of Human3.~6M dataset. For each column (a),(b), and (c), from left to right: 2D ground truth pose in image space, predicted 2D pose by CPN \cite{Chen2018}, 3D ground truth pose in camera space, predicted 3D pose. In ground truth 2D pose and 3D pose, left and right limbs are marked in blue and red respectively, and in the predicted pose, in purple and green respectively.}   
\label{fig:sample_results}
\end{figure*}

\begin{figure}[t]
\begin{center}
%\fbox{\rule{0pt}{2in} \rule{.9\linewidth}{0pt}}
\includegraphics[width=\linewidth]{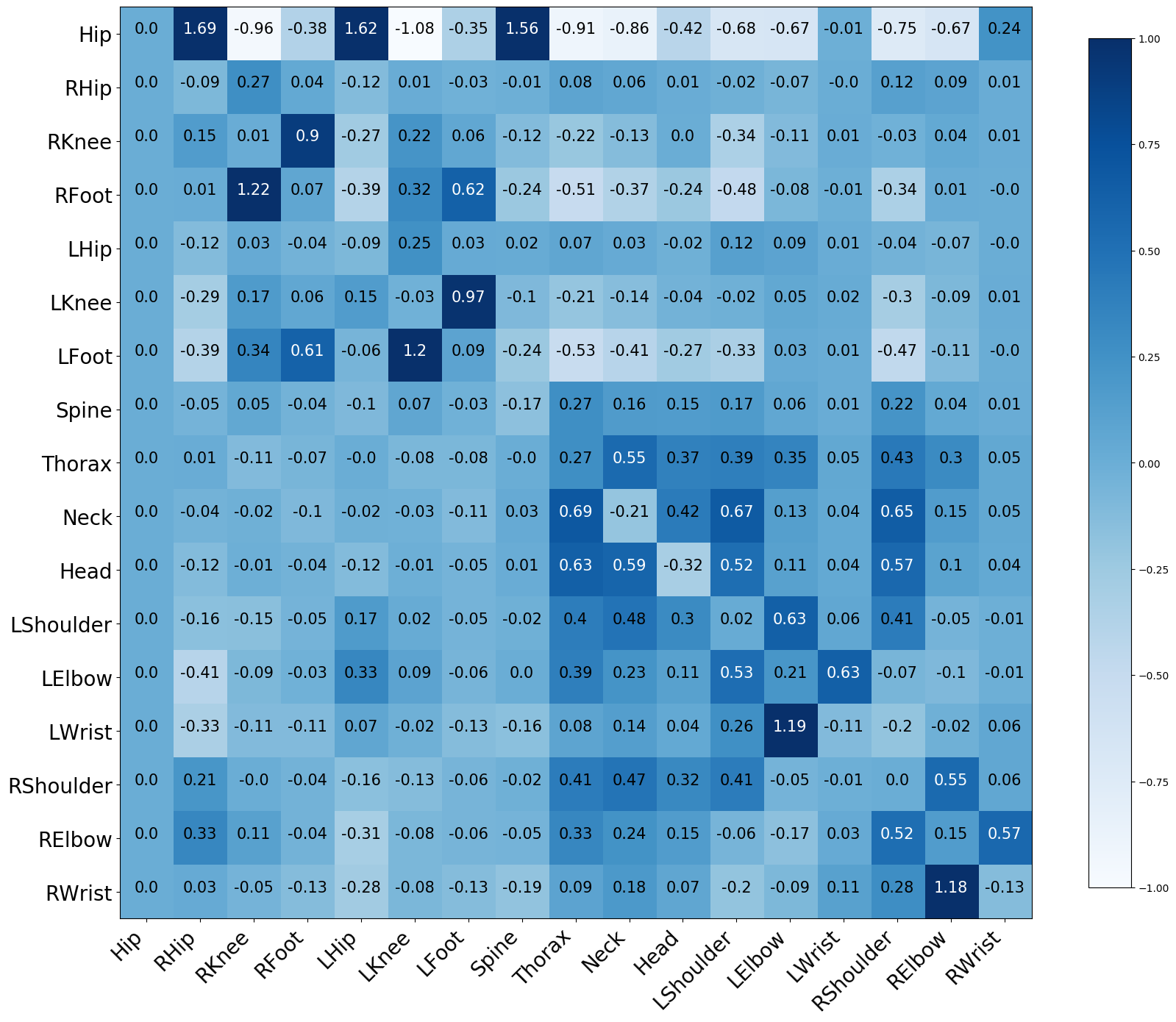}
\end{center}
   \caption{Learned adjaceny matrix $\tilde{A}^{child}$. Dark blue indicates strong relation.}   
\label{fig:adj}
\end{figure}

\subsection{Quantative results}
The average MPJPE score, as well as joint level scores for Protocol \#1, are reported in Table \ref{tab:mpjpe-joint}. 
Most of the error occurs for joints at the body extremities such as head, elbow, foot, and wrist. 
Our model achieves $52.8$ mm of mean error as per Protocol \#1, particularly $17.7, 18.4,$ and $38.9$ mm in x, y, and z axis respectively. 
As can be seen, the majority of the error is in the z or the depth axis.

Table \ref{tab:mpjpe} compares the performance of PoseGraphNet with the state-of-the-art 3D pose estimation methods.
We only consider approaches that operate on a single input frame for this comparison.
The mean MPJPE scores for both protocols are reported in Table \ref{tab:mpjpe}.
The mean scores for individual actions are also reported.
Though our method could not outperform Cai et al.~\cite{Cai2019} in Protocol \#1 and Pavllo et al.~\cite{Pavllo2019} in Protocol \#2, it achieved a better result than Martinez et al.~\cite{Martinez2017}.
Though our model is similar to \cite{Martinez2017}, as we utilize the skeletal structure by using GCN, the performance improves.
We report the performance of our model for both CPN \cite{Chen2018} and Mask-RCNN \cite{he2017mask} predicted inputs.
CPN is a better 2D pose model than Mask-RCNN \cite{Chen2018}, this also translates in the 3D pose prediction results, as can be seen in Table \ref{tab:mpjpe}.

In practice, the 2D to 3D pose regression works in succession to the 2D pose detection model.
So the 2D to 3D model should be minimum in size and have high throughput.
Our motivation for this research is to develop a model that is accurate, fast and small enough to be used in a real-world application such as reFit's Gamo \cite{refit}.
We compare our method with the best performing state-of-the-art method \cite{Cai2019} and also with Martinez et al.'s approach \cite{Martinez2017} in terms of the number of parameters and throughput in Table~\ref{tab:num_param}.
Though our model's performance lags behind the state-of-the-art \cite{Cai2019} by $2.2mm$, it has around $1/3^{rd}$ the number of trainable parameters compared to the other two methods.
We compute the throughput by the number of forward passes in one second with batch size 1.
We compute this in evaluation mode in a machine with GeForce RTX 2060 GPU, AMD Ryzen 9 3900X 12-core processor, and 32GB RAM.
As it is evident from Table \ref{tab:num_param}, our model is accurate enough, even with relatively fewer parameters while maintaining a higher throughput than the state-of-the-art \cite{Cai2019}.

\subsection{Qualitative Results}
Figure \ref{fig:sample_results} shows qualitative results on 9 randomly selected samples from the test set of Human3.6M dataset.
For most of the cases, PoseGraphNet can detect the 3D pose correctly.
It correctly detects the left and right joints, even in cases of the back view of the person (column (b), row 1 from top).
Also, error from the 2D pose detection stage propagates to the 2D to 3D pose part, as can be seen in few examples in Figure~\ref{fig:sample_results} (column (c), row 3).

Figure~\ref{fig:adj} shows the learned adjacency matrix $\tilde{A}^{child}$ of PoseGraphNet.
The adjacency values indicate interesting relationships between the joints.
For example, the right and left feet are strongly related (0.62), though they are not physically connected.
However, the right and left wrists have a weak relation (0.11).
This also follows from intuition, because hands move more independently of each other compared to legs.
It also learns a weak relation (0.21) between the right shoulder and the right hip, suggesting movement of the right shoulder will have a small effect on the right hip.
As we did not enforce any restriction on the adjacency matrix, $\tilde{A}^{child}$ contains connections irrespective of its original initialization of child connections.
%-------------------------------------------------------------------------

\section{Conclusion}
We present a graph convolution network named PoseGraphNet for 3D human pose regression from 2D pose.
Our model is an improved version of the fully connected model proposed in \cite{Martinez2017}. Introduction of the skeletal structural information into the model in the form of the adjacency matrix and graph convolution improves the prediction performance over \cite{Martinez2017}. Our model's performance is close to that of the state-of-the-art \cite{Cai2019}, while having only $1/3^{rd}$ the number of parameters and double the throughput. Additionally, employing an adaptive adjacency matrix allowed us to learn specific adjacency relations between joints, that cannot be handcrafted.

Acknowledgements: The authors acknowledge the financial support by the Federal Ministry of Education and Research of Germany for this project (Start MTI/16SV8115).
%-------------------------------------------------------------------------

\bibliographystyle{IEEEbib}
\bibliography{icip}

\begin{thebibliography}{10}

\bibitem{Moon2018}
Gyeongsik Moon,
\newblock ``{V2V-PoseNet: Voxel-to-Voxel Prediction Network for Accurate 3D
  Hand and Human Pose Estimation from a Single Depth Map},''
\newblock in {\em CVPR}, 2018.

\bibitem{Zimmermann2018}
Christian Zimmermann, Tim Welschehold, Christian Dornhege, Wolfram Burgard, and
  Thomas Brox,
\newblock ``{3D Human Pose Estimation in RGBD Images for Robotic Task
  Learning},''
\newblock in {\em ICRA}, 2018.

\bibitem{Tome2017}
Denis Tome and Chris Russell,
\newblock ``{Lifting from the Deep: Convolutional 3D Pose Estimation from a
  Single Image},''
\newblock in {\em CVPR}, 2017.

\bibitem{Rayat2018}
Mir Rayat, Imtiaz Hossain, and James~J Little,
\newblock ``{Exploiting temporal information for 3D human pose estimation},''
\newblock in {\em ECCV}, 2018.

\bibitem{Cai2019}
Yujun Cai, Liuhao Ge, Jun Liu, Jianfei Cai, Tat-Jen Cham, Junsong Yuan, and
  Nadia~Magnenat Thalmann,
\newblock ``{Exploiting Spatial-temporal Relationships for 3D Pose Estimation
  via Graph Convolutional Networks},''
\newblock in {\em ICCV}, 2019.

\bibitem{Martinez2017}
Julieta Martinez, Rayat Hossain, Javier Romero, and James~J Little,
\newblock ``{A simple yet effective baseline for 3d human pose estimation},''
\newblock in {\em ICCV}, 2017.

\bibitem{marin20183d}
Manuel~J Marin-Jimenez, Francisco~J Romero-Ramirez, Rafael Munoz-Salinas, and
  Rafael Medina-Carnicer,
\newblock ``3d human pose estimation from depth maps using a deep combination
  of poses,''
\newblock {\em Journal of Visual Communication and Image Representation}, 2018.

\bibitem{Zhao2019}
Long Zhao, Xi~Peng, Yu~Tian, Mubbasir Kapadia, and Dimitris~N Metaxas,
\newblock ``{Semantic Graph Convolutional Networks for 3D Human Pose
  Regression},''
\newblock in {\em CVPR}, 2019.

\bibitem{Doosti2020}
Bardia Doosti, Shujon Naha, Majid Mirbagheri, and David~J. Crandall,
\newblock ``{Hope-Net: A graph-based model for hand-Object pose estimation},''
\newblock {\em CVPR}, 2020.

\bibitem{ionescu2013human3}
Catalin Ionescu, Dragos Papava, Vlad Olaru, and Cristian Sminchisescu,
\newblock ``Human3.6m: Large scale datasets and predictive methods for 3d human
  sensing in natural environments,''
\newblock {\em PAMI}, 2013.

\bibitem{Chen2018}
Yilun Chen, Zhicheng Wang, Yuxiang Peng, Zhiqiang Zhang, Gang Yu, and Jian Sun,
\newblock ``{Cascaded Pyramid Network for Multi-person Pose Estimation},''
\newblock {\em CVPR}, 2018.

\bibitem{Pavllo2019}
Dario Pavllo, Eth Z{\"{u}}rich, Christoph Feichtenhofer, David Grangier, Google
  Brain, and Michael Auli,
\newblock ``{3D human pose estimation in video with temporal convolutions and
  semi-supervised training},''
\newblock in {\em CVPR}, 2019.

\bibitem{Pavlakos2018}
Georgios Pavlakos, Xiaowei Zhou, and Kostas Daniilidis,
\newblock ``{Ordinal Depth Supervision for 3D Human Pose Estimation},''
\newblock {\em CVPR}, 2018.

\bibitem{KingmaB14}
Diederik~P. Kingma and Jimmy Ba,
\newblock ``Adam: {A} method for stochastic optimization,''
\newblock in {\em ICLR}, 2015.

\bibitem{glorot2010understanding}
Xavier Glorot and Yoshua Bengio,
\newblock ``Understanding the difficulty of training deep feedforward neural
  networks,''
\newblock in {\em Proceedings of the 13th international conference on
  artificial intelligence and statistics}. JMLR Workshop and Conference
  Proceedings, 2010.

\bibitem{he2017mask}
Kaiming He, Georgia Gkioxari, Piotr Doll{\'a}r, and Ross Girshick,
\newblock ``Mask r-cnn,''
\newblock in {\em Proceedings of the IEEE international conference on computer
  vision}, 2017, pp. 2961--2969.

\bibitem{refit}
{reFit} Systems,
\newblock ``{reFit Gamo - Digital therapy system for individual orthopedic
  rehabilitation},'' 2021.

\end{thebibliography}

\end{document}